\documentclass{article}

\usepackage[numbers]{natbib}
\usepackage[final]{neurips_2022}

\usepackage[utf8]{inputenc} 
\usepackage[T1]{fontenc}    
\usepackage{hyperref}       
\usepackage{url}            
\usepackage{booktabs}       
\usepackage{amsfonts}       
\usepackage{nicefrac}       
\usepackage{microtype}      
\usepackage{xcolor}         
\usepackage{amsmath}
\usepackage{graphicx}
\usepackage{graphics}
\usepackage{subcaption}
\usepackage{accents}
\usepackage{enumitem}
\usepackage{tabu}
\usepackage{wrapfig}
\usepackage{graphicx}
\title{Transcormer: Transformer for Sentence Scoring with Sliding Language Modeling}

\author{%
    Kaitao Song$^{1,}$~\thanks{The first two authors are equal contributions.}, Yichong Leng$^{2,*}$, Xu Tan$^1$, Yicheng Zou$^3$, Tao Qin$^1$, Dongsheng Li$^1$ \\
    Microsoft Research Asia$^1$, University of Science and Technology of China$^2$,  Fudan University$^3$ \\
    \texttt{\{kaitaosong, xuta, taoqin, dongsli\}@microsoft.com},  \\
    \texttt{lyc123go@mail.ustc.edu.cn}, \texttt{yczou18@fudan.edu.cn} 
}
\newcommand{\ie}{\emph{i.e.}}
\newcommand{\eg}{\emph{e.g.}}

\newcommand{\song}[1]{{\color{red} #1}}

\makeatletter
\DeclareRobustCommand{\cev}[1]{%
  \mathpalette\do@cev{#1}%
}
\newcommand{\do@cev}[2]{%
  \fix@cev{#1}{+}%
  \reflectbox{$\m@th#1\vec{\reflectbox{$\fix@cev{#1}{-}\m@th#1#2\fix@cev{#1}{+}$}}$}%
  \fix@cev{#1}{-}%
}
\newcommand{\fix@cev}[2]{%
  \ifx#1\displaystyle
    \mkern#23mu
  \else
    \ifx#1\textstyle
      \mkern#23mu
    \else
      \ifx#1\scriptstyle
        \mkern#22mu
      \else
        \mkern#22mu
      \fi
    \fi
  \fi
}

\makeatother

\begin{document}

\maketitle

\begin{abstract}
    Sentence scoring aims at measuring the likelihood score of a sentence and is widely used in natural language processing scenarios, like reranking, which is to select the best sentence from multiple candidates. Previous works on sentence scoring mainly adopted either causal language modeling (CLM) like GPT or masked language modeling (MLM) like BERT, which have some limitations: 1) CLM only utilizes unidirectional information for the probability estimation of a sentence without considering bidirectional context, which affects the scoring quality; 2) MLM can only estimate the probability of partial tokens at a time and thus requires multiple forward passes to estimate the probability of the whole sentence, which incurs large computation and time cost. In this paper, we propose \textit{Transcormer} -- a Transformer model with a novel \textit{sliding language modeling} (SLM) for sentence scoring. Specifically, our SLM adopts a triple-stream self-attention mechanism to estimate the probability of all tokens in a sentence with bidirectional context and only requires a single forward pass. SLM can avoid the limitations of CLM (only unidirectional context) and MLM (multiple forward passes) and inherit their advantages, and thus achieve high effectiveness and efficiency in scoring. Experimental results on multiple tasks demonstrate that our method achieves better performance than other language models. Our code and pre-trained models will be released at: \url{https://github.com/microsoft/CyBERTron-LM/Transcormer}. 
\end{abstract}

\section{Introduction}

Sentence scoring is to measure the log-likelihood score of a sentence via language model, so that it can be used to represent the  relative likeliness of a sentence. Specifically, a good sentence should have a relatively lower log-likelihood score, which means more linguistically acceptable for the sentence~\cite{Jey2017Grammaticality}.  Due to such nature, sentence scoring has been widely used in many natural language processing (NLP) scenarios. For instance, it can be used to rerank candidates in neural machine translation (NMT) or automatic speech recognition (ASR) tasks~\cite{Bengio2015nmtrerank,William2016rerankasr} or evaluate sentences in linguistic acceptability~\cite{Alex2020BLiMP}. Therefore, how to design effective language models to calculate sentence scores efficiently is very important.

Recently, neural network based language models (LM)~\cite{Nathan2019WMT,Yingce2019WMT,Yee2019Noisy,Yuchen2018Reranking} have been considered as the most widely used technique for sentence scoring, since they can produce density estimation of the whole sentence by computing the probability of each token and summing up their values as the sentence score. Specifically, causal language modeling (CLM)~\cite{Radford2018GPT} and masked language modeling (MLM)~\cite{devlin2019bert} are the most representative LMs. For a given sentence ${\bf x} = \{x_1, \cdots, x_{|\bf{x}|}\}$, where $x_i$ is the $i$-th token of ${\bf x}$. CLM 
can predict next token conditioned on unidirectional context and its objective is to optimize $\sum_{i=1}^{|\bf{x}|} \log P(x_i|x_{<i})$. Hence, CLM is usually used for solving natural language generation (NLG) tasks~\cite{Radford2018GPT,radford2019GPT2,Tom2020GPT3}. While for MLM, it replaces a subset of tokens $\bf{x}_{\cal{S}}$ in $\bf{x}$ as special symbol $\rm{[MASK]}$ and then predicts the masked tokens based on the corrupted sequence $\bf{x}_{\backslash \cal{S}}$. The objective of MLM is to optimize $\sum_{i=1}^{|\cal{S}|} \log P(x_{{\cal S}_i}|{\bf x}_{\backslash \cal{S}})$, so it is able to learn bidirectional context and can be used for solving natural language understanding (NLU) tasks~\cite{devlin2019bert,Yinhan2019Roberta,Zhenzhong2020Albert,Dong2019UniLM}. Since CLM and MLM can be learned in an unsupervised fashion, many works ~\cite{devlin2019bert,Yinhan2019Roberta,Zhenzhong2020Albert,Radford2018GPT,radford2019GPT2,Tom2020GPT3,Yang2019XLNet,Song2020MPNet,song2019mass,Dong2019UniLM,Mike2020BART,raffel2019exploring} have pre-trained these LMs on large-scale corpus to extract powerful linguistic representations, and these models can be directly used out of the box to predict the probabilities of tokens. Inspired by the successes of LMs, some works ~\cite{Julian2020MLMScoring,Alex2019BERTMouth,Joonbo2019reranking,Chiu2021Reranking,Yue2020GPTRerank} have tried to use pre-trained CLM or MLM to generate sentence scores on NMT or ASR reranking tasks and achieved some promising results. 

However, we notice that both CLM and MLM still suffer from some deficiencies in calculating sentence scores. For instance, some works~\cite{Yue2020GPTRerank} utilized GPT-style model for ASR reranking within single inference, yet GPT model can only extract unidirectional information due to the limitation of CLM, without considering the whole sentence semantics, and thus affect the sentence score. To utilize bidirectional context, some works~\cite{Alex2019BERTMouth,Joonbo2019reranking,Chiu2021Reranking} applied BERT model for rescoring. However, the nature of MLM is to mask some tokens in the sentence for prediction, which means it requires the BERT model to forward multiple times and each forward pass only masks one token for prediction. As a result, it is time-consuming to directly adopt MLM for scoring sentence. In summary, we find that for calculating sentence scores, CLM needs one-pass inference but only uses unidirectional information and MLM is costly in computing sentence score although it can use bidirectional context. Therefore, a natural question arises: is it possible to design a language model to use bidirectional context for sentence scoring with only one inference pass. 

In this paper, we introduce \textit{Transcormer}, a Transformer model designed for sentence scoring. More specifically, our Transcormer leverages a novel language modeling scheme, named as \textit{sliding language modeling} (SLM), that  produces the probability of all tokens within single inference pass and simultaneously utilizes bidirectional context. To fulfill this target, we innovatively design a triple-stream self-attention mechanism, which consists of two content streams (a forward stream and a backward stream) and one query stream. By employing  specifically-designed mask strategies on the attention matrix, our method allows each token in the query stream to leverage all token information except itself (i.e., the tokens before and after it) for estimating its probability to avoid any information leakage. 
To the best of our knowledge, SLM is the first language modeling tailored for sentence scoring. We pre-train our SLM on large-scale corpus, and then evaluate it on multiple datasets. Experimental results demonstrate that Transcormer outperforms the baselines by up to + 0.8/0.6 BLEU score on small/large-scale NMT tasks, $\sim$20\% relative improvements on ASR tasks.

The main contributions of this work are summarized as follows:
\begin{itemize}[leftmargin=*]
    \item We analyze the pros and cons of CLM and MLM when using them for scoring sentences, and propose Transcormer with a new sliding language modeling, which uses bidirectional context for probability estimation within a single pass.
    \item We introduce a novel triple-stream self-attention mechanism in SLM, which has two content streams to collect forward/backward semantics, and a query stream to estimate the probability of each token in a sentence.
    \item Experimental results on multiple datasets demonstrate the effectiveness and efficiency of our SLM for sentence scoring.
\end{itemize}

\section{Background}
\label{sec2:background}

\subsection{Sentence Scoring}
\label{sec:sentence_scoring}
Sentence scoring has a long history in NLP applications, especially in reranking tasks (\eg, reranking for machine translation~\cite{Shankar2004SMT,Raphael2017SMT,Libin2004SMT,Franz2003SMT} or speech recognition~\cite{Yukun2017ASR}). Generally, given $n$-best candidates generated by text generation models, reranking aims at scoring each candidate to select the best answer. Early works~\cite{Shankar2004SMT,Raphael2017SMT,Libin2004SMT,Franz2003SMT,Auli2014Rerank,Terumasa2018SMT}  mainly used statistical LM or combined it with RNN-based LM to calculate the sentence scores. Recently, end-to-end neural network based LM has became the de facto approach for scoring and has been applied in many NLP tasks~\cite{Yingce2019WMT,Nathan2019WMT,Yuguang2017WMT}. Specifically, causal language modeling (CLM) and masked language modeling (MLM) are the most representative language modeling methods, among which GPT~\cite{Radford2018GPT,radford2019GPT2,Tom2020GPT3} and BERT~\cite{devlin2019bert} are the most famous examples, respectively. As aforementioned, CLM conditions on the previous states to predict next token so that it can obtain the probability of each token of the sentence in a single pass, but can only capture unidirectional information. For MLM, it is able to use bidirectional context for prediction but the masked-prediction mechanism limits MLM to producing the probability of all tokens within one forward pass (since MLM can only provide the probability of masked tokens). To calculate the sentence score, a kind of solutions~\cite{Alex2019BERTMouth,Joonbo2019reranking,Chiu2021Reranking} is to forward multiple times and only mask one token each time. Therefore, the MLM for sentence scoring is formulated as: $\sum_{i=1}^{|{\bf x}|} \log P(x_i|{\bf x}_{\backslash x_i})$. We can find that the cost of MLM for scoring needs $|{\bf x}|$ inference passes, which is too time-consuming. 

Some recent works~\cite{Alex2019BERTMouth,Julian2020MLMScoring,Kevin2020Cloze,Joongbo2020DLM} have been proposed to alleviate these issues in MLM. Wang et al.~\cite{Alex2019BERTMouth} and Salazar et al.~\cite{Julian2020MLMScoring} attempted to use stochastic estimation or distillation to avoid N-passes problem to approximately estimate the probability of each toke produced by MLM, with a sacrifice of performance. Clark et al. \cite{Kevin2020Cloze} adopted a two-cloze tower~\cite{Alexei2019Cloze} with noise-contrastive estimation to provide sentence probability, and Shin et al. \cite{Joongbo2020DLM} only considered word embedding as the inputs of key and value in transformer without any interaction. Besides, some works~\cite{Sumanta2021Energy,Lee2021Discriminative} adopted discriminative language modeling to approximately estimate sentence scores based on the paired data, but this paradigm must require labeled data from the downstream tasks and cannot utilize unlabled data for pre-training. More discussions about related works can be found in Appendix~\ref{appendix:related_works}. Therefore, how to calculate sentence scores efficiently with bidirectional context is the main challenge in MLM.

Overall, CLM only needs a single forward pass to estimate the probability of all tokens but cannot extract bidirectional context, while MLM leverages bidirectional information but needs multiple inference passes. Consequently, we raise a natural question: is it possible to design a pre-trained language model to support all token prediction in a single pass and simultaneously leverage bidirectional information? This is exactly the motivation of our method.

\subsection{Multiple-Stream Self-Attention}

The pioneer of multiple-stream self-attention is XLNet~\cite{Yang2019XLNet}, which introduces two-streams self-attention to incorporate autoregressive pre-training for language understanding, which pre-trains Transformer~\cite{Vaswani2017Transformer} via using a content stream and a query stream. 
In details, for the $t$-th step, the content stream is able to capture the dependency from the tokens before the $t$-th step and itself (\ie, $x_{\le t}$), while the query stream is only allowed to view the tokens before the $t$-th step (\ie, $x_{< t}$) to avoid information leakage. Besides, there are some other variants of two-stream self-attention~\cite{Song2020MPNet,Weizhen2020ProphetNet,Dongling2020Erinegan}, which are designed for solving NLU tasks. For example, MPNet~\cite{Song2020MPNet} used two-stream self-attention to build masked and permuted pre-training. ProphetNet~\cite{Weizhen2020ProphetNet} designed multiple query streams to predict N-gram future steps for sequence-to-sequence tasks, and ERNIE-GAN~\cite{Dongling2020Erinegan} proposed a multi-flow generation model, which includes two query streams for span and word prediction. We observe that these works mainly used a single content stream and then used one or many query streams to predict more information. Different from these works, we introduce a triple-stream self-attention mechanism, which enables query stream to leverage two content streams for prediction, and thus enjoys the benefits of additional bidirectional context for estimating token probability. %

\section{Transcormer}
\label{sec3:method}
\begin{figure}
    \centering
    \includegraphics[width=0.95\textwidth]{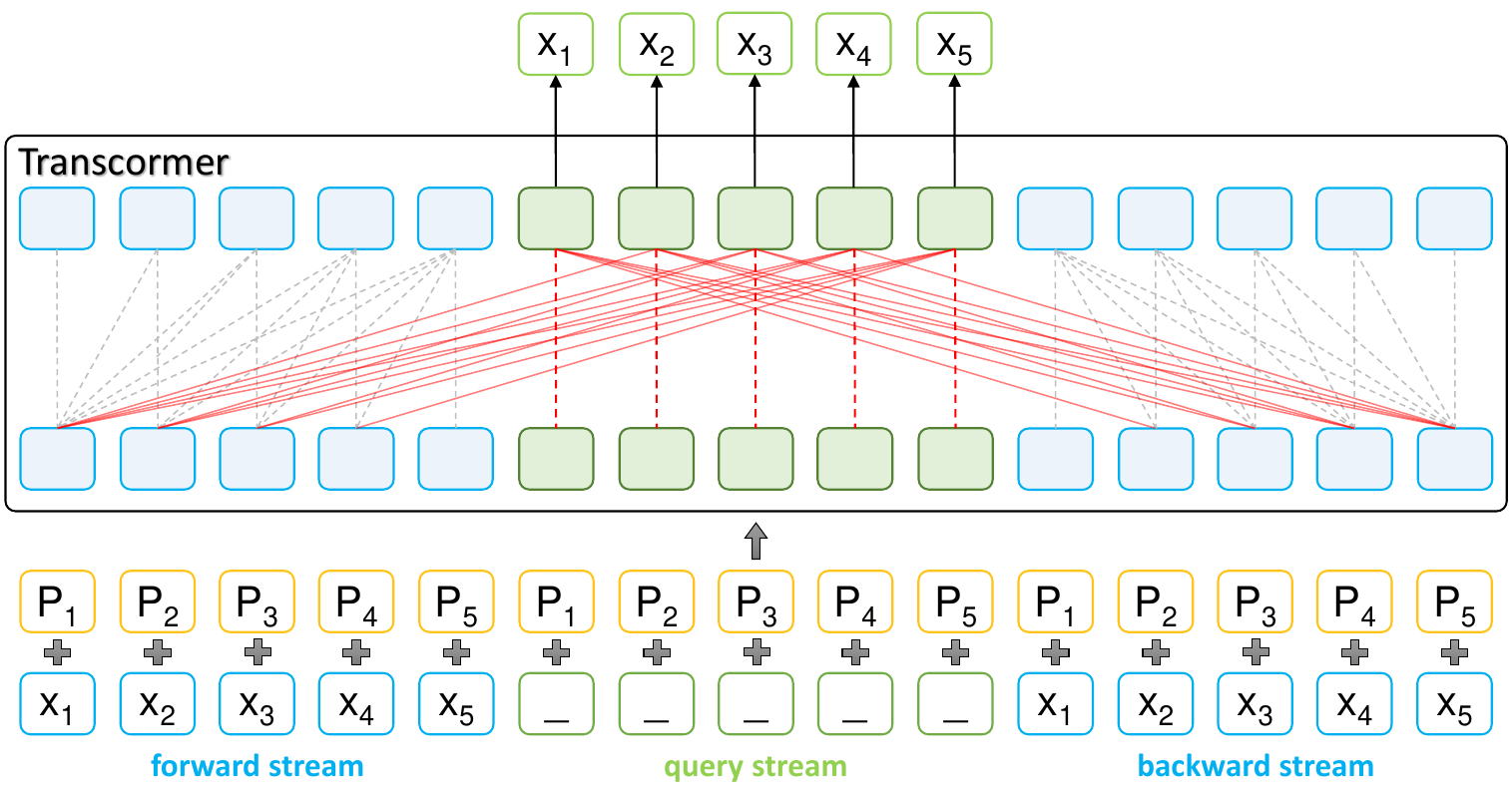}
    \vspace{-5pt}
    \caption{Transcormer with sliding language modeling. The left and the right (in blue) are forward and backward streams, respectively, and the middle (in green) is query stream. For query stream, the inputs are only the positional information. We use gray and red line to represent the allowed attended positions in the content and query streams.}
    \label{fig:slm}
\end{figure}

To inherent the advantages of CLM and MLM for sentence scoring and avoid their limitations, we propose Transcormer -- a Transformer model with a novel sliding language modeling for sentence scoring. First, we summarize that an ideal language modeling for sentence scoring should satisfy two requirements: 1) model should be able to use bidirectional context for effective probability estimation of each token; 2) model should produce the probability of all tokens in a sentence within a single inference pass for efficiency. Therefore, to fulfill these two requirements, we formulate a new language modeling, named as sliding language modeling (SLM), and describe it in Section~\ref{subsec:slm}. In SLM, we propose a Triple-Stream Self-Attention mechanism based on Transformer (please see Section~\ref{subsec:tssa} for details) to use bidirectional context for each token prediction and avoid information leakage. We also discuss the differences between SLM and other LMs in Section~\ref{subsec:discuss}. Figure~\ref{fig:slm} presents the pipeline of our Transcormer for sentence scoring with SLM.

\subsection{Sliding Language Modeling}
\label{subsec:slm}
Considering the pros and cons of CLM and MLM for scoring, we notice: 1) CLM can produce probability of all tokens within one forward pass,  and thus obtain the unidirectional information of the whole sentence; 2) MLM for sentence scoring needs multiple inference passes, so that many context has been repeatedly calculated and cause a waste of computation. So, is it possible to reuse token information to build bidirectional context for token prediction?

Therefore, we propose sliding language modeling (SLM) to address the inherent flaws in previous LMs (\ie, CLM and MLM) for sentence scoring. Specifically, we first maintain two individual streams to collect forward (left-to-right) context and backward (right-to-left) context. And for each token prediction, we decompose the sentence information as the past tokens (the tokens before it) and future tokens (the tokens after it) respectively. As a result, our SLM enforces each token to only capture the dependency from its past tokens and its future tokens concurrently, so that each token can utilize the whole sentence information (except itself) to estimate token probability. The objective function of SLM is formulated as:
\begin{equation}
    \label{eq2}
    {\cal L} = \sum_{i=1}^{|{\bf x}|} \log P(x_i| {\bf x}_{<i}, {\bf x}_{>i}; \theta),
\end{equation}
where ${\bf x}_{<i}$ and ${\bf x}_{>i}$ respectively correspond to the tokens before the $i$-th token and after the $i$-th token, and $\theta$ represents the parameters of SLM. Thanks to such design, our SLM can utilize bidirectional context to produce the probability of each token within one forward pass, and thus satisfy the above requirements for sentence scoring. However, previous experiences~\cite{Yang2019XLNet,Jungo2020DisCo} pointed out that the states with bidirectional information will cause information leakage when propagating to the next layer~\footnote{For example, assume the sequence has 3 tokens, and the hidden states of the $i$-th token at the first layer as $h_{i}^1$. So each position $h_{1}^1$, $h_{2}^1$ and $h_{3}^1$ should collect the information from positions $\{2,3\}$, $\{1,3\}$, $\{1,2\}$. However, when $h_{1}^1$ and $h_{3}^1$ are delivered to $h_{2}^2$ at the second layer, it will cause a cyclic leakage as $h_{2}^2$ should not obtain information from position 2.}. Therefore, how to implement SLM to avoid information leakage and maintain different states together is still a troublesome problem. 

\begin{figure*}
    \centering
    \begin{subfigure}[t]{0.56\textwidth}
        \centering
        \includegraphics[width=\textwidth]{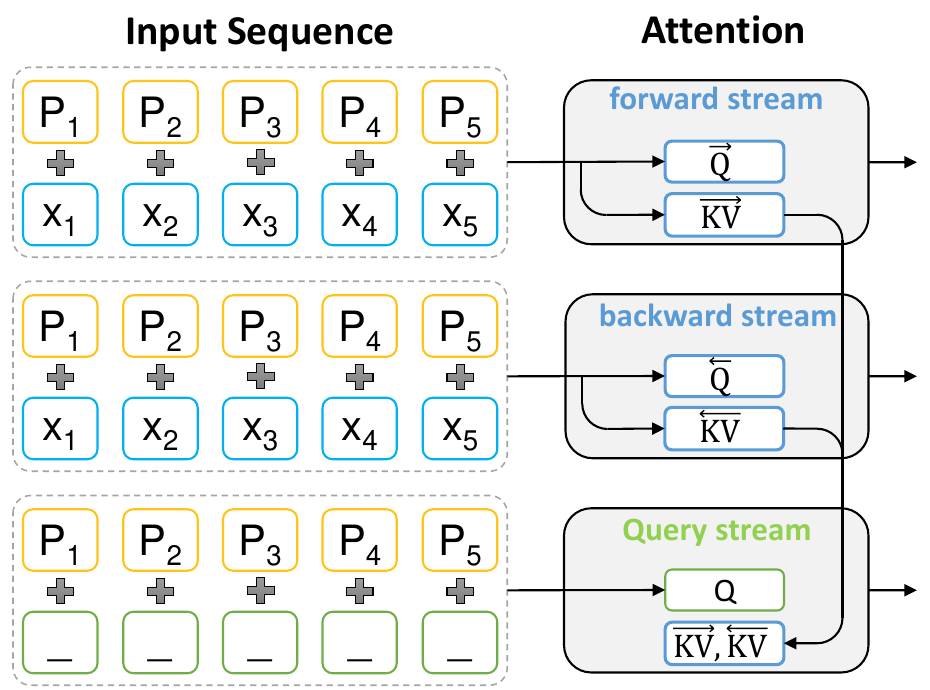}
        \caption{}
        \label{MPNet_a}
    \end{subfigure}
    \begin{subfigure}[t]{0.42\textwidth}
        \centering
        \includegraphics[width=\textwidth]{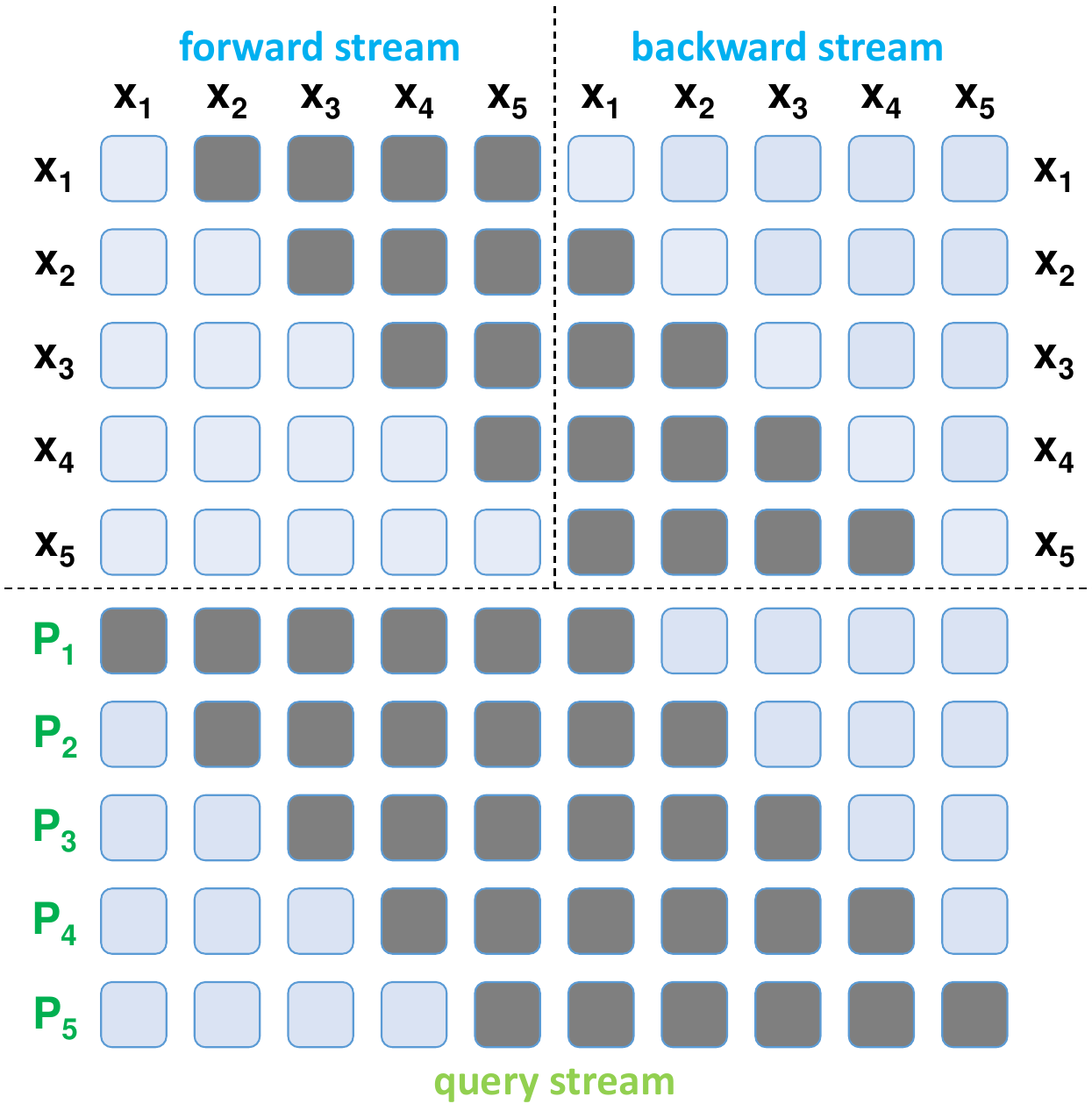}
        \caption{}
        \label{MPNet_a}
    \end{subfigure}
    \caption{(a) The structure of our triple-stream self-attention used in our sliding language modeling. The query stream reuses the hidden states from both forward and backward (content) stream as the key and value in attention. (b) The attention mask matrix used in our triple-stream self-attention. The above row is the attention matrix for the forward and backward stream and the below row is the attention matrix for the query stream. The cell in gray color means this position cannot be attended.}
    \label{fig:tssa}
\end{figure*}
\subsection{Triple-Stream Self-Attention}
\label{subsec:tssa}
Based on the Eqn~\ref{eq2} of our proposed SLM, model needs to maintain two states to collect forward and backward contexts for prediction, and we call these two states as the forward stream and backward stream respectively. To avoid information leakage, we additionally maintain an individual state for prediction and control it to only capture the dependency from the forward and backward streams. We name this state as the query stream. Therefore, we propose a novel Triple-Stream Self-Attention to update each stream, and the detailed design is described as following.

To fulfill our target, we choose Transformer~\cite{Vaswani2017Transformer} as our basic model, due to its flexibility in capturing global dependency. Assume the input sequence as $\{w_1, w_2, \cdots, w_n\}$ and its positions as $\{p_1, p_2, \cdots, p_n\}$, where $w_i$ and $p_i$ represents the embedding of $i$-th token and its position in the sentence, and $n$ is the token number. For the query stream, we only use the positional embeddings $\{p_1, p_2, \cdots, p_n\}$ as the input. For the forward and backward streams, we also maintain two individual states and both of them use the tokens plus its position (\ie, $\{w_1 + p_1, w_2 + p_2, \cdots, w_n + p_n\}$) as the input. 
For the $l$-th layer calculation, we denote the forward and backward streams of the position $i$ as $\vec{h}_{i}^l$ and $\cev{h}_{i}^l$, and they are updated as:
\begin{align}
    \vec{h}_{i}^l &= {\rm Attention} ({\bf Q} = \vec{h}_{i}^{l-1}, {\bf KV} = \vec{h}_{\song{\le i}}^{l-1}; \theta), \\
    \cev{h}_{i}^l &= {\rm Attention} ({\bf Q} = \cev{h}_{i}^{l-1}, {\bf KV} = \cev{h}_{\song{\ge i}}^{l-1}; \theta),
\end{align}
where $\rm{Attention(\cdot, \cdot)}$ refers to the self-attention~\cite{Vaswani2017Transformer} in Transformer and \textbf{Q}, \textbf{K}, \textbf{V} denote the query, key and value in self-attention. Hence, $\vec{h}_{i}^l$ collects the information before the position $i$ and itself, and $\cev{h}_{i}^l$ collects the information after the position $i$ and itself. For the query stream, we denote $q_i^l$ as its hidden states, and concatenate forward stream $\vec{h}_{i}^l$ and backward stream $\cev{h}_{i}^l$  of the current layer to be the key/value of query stream. So $q_i^l$ is updated as:
\begin{equation}
    \label{eq5}
   q_{i}^l = {\rm Attention} ({\bf Q} = q_{i}^{l-1}, {\bf KV} = \left[ \vec{h}_{\song{< i}}^{l-1}, \cev{h}_{\song{> i}}^{l-1} \right]; \theta).
\end{equation}
Here we find that $q_i^l$ is required to only capture the dependency from the forward stream before the position $i$ and the backward stream after the position $i$, rather than itself. Due to such design, the query stream is able to capture bidirectional context for estimating token probability and avoid information leakage, which is more effective than CLM in using context. More importantly, our triple-stream self-attention enables model to predict the probability of all tokens in a sentence within a single forward pass, which demonstrates more efficiency than MLM. Figure~\ref{fig:tssa} presents the detailed design of our triple-stream self-attention. {We can find that in the query stream, the masked matrix is like a sliding window to control each token to view its previous states in forward stream and its future states in backward stream. And that is why we name our model as sliding language modeling.} 

\begin{table}[!t]
    \centering
    \begin{tabular}{l| l | l | r | r}
    \toprule
    LM     &  Model & Cost & Context & Scenario \\
    \midrule
    CLM & GPT~\cite{radford2019GPT2} & $\times 1$ & forward & NLG \\
    MLM & BERT~\cite{devlin2019bert} & $\times {\rm n}$  & bidirectional & NLU \\
    Bi-LM & ELMO~\cite{Matthew2018ELMO} & $\times 2$  & forward + backward & NLU \\
     \midrule
    SLM     &  Transcormer & $\times 3$ & bidirectional & Scoring \\
    \bottomrule
    \end{tabular}
    \vspace{5pt}
    \caption{Comparisons between SLM and other LMs. We assume all LMs adopt the same architectures (\eg, Transformer). The ``Cost'' column means the relative computations compared with CLM when calculating a sentence with $n$ tokens. The ``{Context}'' column means the contextual information usage for prediction. }
    \label{tab:comparison}
    
\end{table}

\subsection{Discussion}
\label{subsec:discuss}
To better understand our SLM, we analyze the advantages of our SLM over other LMs. The comparisons are listed in Table~\ref{tab:comparison}. We select three representative LMs for comparisons, which are CLM (BERT), MLM (GPT) and bidirectional LM (Bi-LM, used in ELMO~\cite{Matthew2018ELMO}~\footnote{ELMO pre-trains a left-to-right and a right-to-left LSTM~\cite{Sepp1997LSTM} and concatenates the outputs of each last unidirectional LSTM layer for prediction.}) respectively. From Table~\ref{tab:comparison}, we have the following observations: 
\begin{enumerate}[leftmargin=*]
    \item When compared with CLM, our SLM requires 3 $\times$ computations. However, our SLM can fully use the whole sentence information for prediction while CLM can only use unidirectional information. Even scaling CLM as 3 $\times$ parameters, it still can not use bidirectional context for prediction. This also demonstrates the effectiveness of our SLM in using context.
    
    \item  MLM is powerful at extracting bidirectional context but it needs $n \times$ inferences to calculate the whole sentence information limited to its masked prediction. Our SLM just needs a single inference and uses bidirectional information for prediction with only $3 \times$ computations. Especially, our SLM shows higher efficiency compared with MLM when $n$ is  large.
    
    \item Bi-LM can also extract forward and backward contextual information, but it just simply concatenates the forward and backward features for the final prediction, without any interactions. Instead, our SLM can iteratively fuse the bidirectional information thanks to our triple-stream self-attention mechanism.
\end{enumerate}

Overall, the design of SLM is dedicated for sentence scoring, while CLM prefers NLG tasks and MLM/Bi-LM prefer NLU tasks.

\section{Experiments}
In this section, we describe our experimental setup, and the results on NMT and ASR datasets.
\label{sec4:exp}
\subsection{Experimental Setup}
\label{subsec:exp_setup}

We adopt Transformer~\cite{Vaswani2017Transformer} as the backbone network. Following previous works~\cite{devlin2019bert}, we adopt a base setting and a small setting for our model as Transcormer$_{base}$ with 110M parameters and Transcormer$_{small}$ with 34M parameters, that consists of 12/6 transformer layers and each layer has 768/512 hidden size and 12/8 attention heads. During the pre-training, we use wikipedia plus bookcorpus (16GB) as the training corpus, to be consistent with previous works~\cite{devlin2019bert}. Our model is trained at the sentence-level (\ie, one sentence per sample). We choose Adam~\cite{Diederik2015Adam} as the default optimizer with learning rate of $5e-4$, $\beta_1 = 0.9$, $\beta_2 = 0.98$ and $\epsilon = 1e-6$, and weight decay is set as 0.01. The learning rate warms up over the first 10,000 steps and then linearly decays. We set the batch size as 8192 tokens per batch, and the training step is 125,000 steps. We use 32 NVIDIA Tesla 32GB GPUs, with FP16 speedup. The total training needs 5.5 days for Transcormer$_{base}$. For more experimental settings (\eg, dataset and its split sizes), please refer to the Appendix. The code and pre-trained models will be released at: \url{https://github.com/microsoft/CyBERTron-LM/Transcormer}. 

\begin{table}[!t]
    \small
    \centering
    \begin{tabular}{l |c c c c c c c c | c}
    \toprule
              & \multicolumn{8}{c|}{IWSLT} & WMT \\
    Model     & De & Es & It & Nl & Pl & Ro & Ru & Tr & De-En \\
    \midrule
    Oracle & 41.80 & 48.69 & 41.89 & 44.38 & 27.90 & 46.01 & 29.60 & 27.25 & 39.17 \\
    \midrule
    Baseline  & 34.77 & 41.20 & 34.95 & 37.73 & 22.67 & 38.73 & 24.21 & 21.65 & 32.54 \\
    CLM (GPT) & 34.96 & 41.39 & 35.14 & 38.08 &  22.91 & 39.03 & 24.62 & 22.14 & 32.88 \\
    MLM (BERT) & 35.14 & 41.54 & \textbf{35.54} & 38.14 &  23.00 & 39.21 & 24.65 & 22.36 & 33.07 \\
    Bi-LM (ELMO) & 35.10 & 41.52 & 35.21 & 38.03 &  23.09 & 39.07 & 24.53 & 21.91 & 32.90 \\
    \midrule
    SLM (Transcormer$_{base}$) & \textbf{35.24} & \textbf{41.86} & 35.52 & \textbf{38.45} & \textbf{23.29} & \textbf{39.34} & \textbf{24.69} & \textbf{22.41} & \textbf{33.10} \\
    SLM (Transcormer$_{small}$) & 35.05 & 41.58 & 35.15 & 38.06 &  22.98 & 39.08 & 24.52 & 22.06 & 32.94 \\
    \bottomrule
    \end{tabular}
    
    \vspace{5pt}
    \caption{Reranking results on IWSLT and WMT tasks, and all LMs have the same model architecture as Transcormer. The translation direction of all IWSLT tasks is to English and all results are reported in BLEU score. All LMs are pre-trained in the wikipedia + bookcorpus (16GB) with the same optimization. The last row is the oracle score from the generated candidates.}
    \label{tab:nmt}
\end{table}

\subsection{Experiments on Neural Machine Translation}
\label{subsec:nmt}
We choose IWSLT14 dataset~\cite{cettolo2014report}, which includes multiple small-scale translation tasks from different languages to English, and WMT14 English-German dataset~\footnote{Here we only evaluate German$\rightarrow$English direction as our model is trained on English domain.} for evaluation. We adopt Transformer~\cite{Vaswani2017Transformer} as the machine translation model with 6-6 transformer layers to generate multiple candidates for reranking, with a beam size of 10. The hidden size and attention head are set as 512/1024 and 8/16 for IWSLT and WMT tasks respectively. {During the reranking, we combine the original score produced by the machine translation model and the LM score with a hyper-parameter $\lambda$, by following previous experiences~\cite{Yee2019Noisy,Julian2020MLMScoring}.} The hyper-parameter $\lambda$ is tuned on dev set with a range of $[0.0, 2.0]$, and then select the best $\lambda$ to evaluate the test set. The results are reported in Table~\ref{tab:nmt}, in terms of BLEU~\cite{Kishore2002Bleu}. Each task will be evaluated by five times based on different pre-trained checkpoints, and report the mean value, with a variance of 0.05.
From Table~\ref{tab:nmt}, we have the following observations: 
\begin{itemize}[leftmargin=*]
    \item Our Transcormer$_{base}$ can obtain better performance than CLM and Bi-LM~\footnote{Here, we replace LSTM as Transformer in Bi-LM to keep the consistence in architecture.} in both small-scale IWSLT and large-scale WMT tasks, which indicates the importance of bidirectional context for sentence scoring, and further validate the ability of our SLM in utilizing bidirectional information.
    \item When compared with MLM, our Transcormer$_{base}$ also achieves comparable results. Considering that the computation of MLM for scoring is linear to the input length and needs $n$ inference passes, our SLM show higher efficiency with only $3 \times$ computations in a single pass to maintain three streams. 
\end{itemize}

{Overall, all of the results reveal that our model is more effective in using contextual information for probability estimation and more efficient with only a single forward pass, especially for long sentences.}

\begin{table}[h]
    \centering
    \begin{tabular}{l |c c c c }
    \toprule
    Model     & dev-clean & dev-other & test-clean & test-other \\
    \midrule
    Baseline  & 2.80 & 6.90 & 3.06 & 7.05 \\
    CLM (GPT) & 2.47 & 6.13 & 2.73 & 6.33\\
    MLM (BERT) & 2.30 & 5.65 & 2.59 & 5.90 \\
    Bi-LM (ELMO) & 2.41 & 5.92 & 2.63 & 6.12 \\
    \midrule
    SLM (Transcormer$_{base}$) & \textbf{2.23} & \textbf{5.54} & \textbf{2.49} & \textbf{5.72} \\
    
    SLM (Transcormer$_{small}$) & 2.48 & 5.95 & 2.62 & 6.20 \\
    \midrule 
    Oracle & 1.45 & 4.23 & 1.59 & 4.19\\
    \bottomrule
    \end{tabular}
    \vspace{5pt}
    \caption{Reranking results on LibrSpeech dataset. All results are reported in WER.}
    \label{tab:asr}
\end{table}

\subsection{Experiments on Automatic Speech Recognition}
We choose LibrSpeech~\cite{Vassil2015Librispeech} to evaluate the performance of our model for reranking on ASR task. We train a Conformer model \cite{gulati2020conformer} on LibrSpeech, which has 12 encoder layers and 6 decoder layers with 512 hidden size and 8 attention heads, and the beam size is set as 10. In addition, we use SpecAug \cite{park2019specaugment} as a data augmentation technology to further improve the accuracy of ASR system. We use word error rate (WER) to evaluate the performance of ASR tasks. We follow the same tuning technique used in NMT tasks for hyper-parameter $\lambda$, but with a larger range as $[0.0, 5.0]$. The results are reported in Table~\ref{tab:asr}. From Table~\ref{tab:asr}, we find that our model can give nearly 20\% relative improvements over the baseline and also outperform other LMs, including CLM, MLM and Bi-LM. The results on ASR task further demonstrates the generalization and effectiveness of our SLM in sentence scoring.  

\newcommand\rotation{30}
\begin{table*}[!t]
    \begin{minipage}{1.0\linewidth}
	    \centering
	    \footnotesize
	    \addtolength{\tabcolsep}{-3.8pt}

        \begin{tabu}{@{}ll@{\hskip 8pt}llllllllllll@{}}
            \multicolumn{1}{p{2.5ex}}{\rotatebox{0}{\textbf{Model}}} &
            \multicolumn{1}{p{2.5ex}}{\rotatebox{\rotation}{\textbf{Overall}}} &
            \multicolumn{1}{p{2.5ex}}{\rotatebox{\rotation}{\textsc{Ana.\ agr}}} &
            \multicolumn{1}{p{2.5ex}}{\rotatebox{\rotation}{\textsc{Arg.\ str}}} &
            \multicolumn{1}{p{2.5ex}}{\rotatebox{\rotation}{\textsc{Binding}}} &
            \multicolumn{1}{p{2.5ex}}{\rotatebox{\rotation}{\textsc{Ctrl.\ rais.}}} &
            \multicolumn{1}{p{2.5ex}}{\rotatebox{\rotation}{\textsc{D-n agr}}} &
            \multicolumn{1}{p{2.5ex}}{\rotatebox{\rotation}{\textsc{Ellipsis}}} &
            \multicolumn{1}{p{2.5ex}}{\rotatebox{\rotation}{\textsc{Filler\ gap}}} &
            \multicolumn{1}{p{2.5ex}}{\rotatebox{\rotation}{\textsc{Irregular}}} &
            \multicolumn{1}{p{2.5ex}}{\rotatebox{\rotation}{\textsc{Island}}} &
            \multicolumn{1}{p{2.5ex}}{\rotatebox{\rotation}{\textsc{NPI}}} &
            \multicolumn{1}{p{2.5ex}}{\rotatebox{\rotation}{\textsc{Quantifiers}}} &
            \multicolumn{1}{p{2.5ex}}{\rotatebox{\rotation}{\textsc{S-v agr}}} \\
            \toprule 
            GPT-2 (345M) & 82.6 & \textbf{99.4} & 83.4 & 77.8 & 83.0 & 96.3 & 86.3 & 81.3 & 94.9 & 71.7 & 74.7 & \textbf{74.1} & 88.3 \\
            \midrule
            BERT (base) & 84.2 & 97.0 & 80.0 & {82.3} & 79.6 & \textbf{97.6} & 89.4 & 83.1 & 96.5 & 73.6 & 84.7 & 71.2 & \textbf{92.4} \\
            BERT (large) & 84.8 & 97.2 & 80.7 & 82.0 & 82.7 & \textbf{97.6} & 86.4 & 84.3 & 92.8 & 77.0 & 83.4 & 72.8 & 91.9  \\
            RoBERTa (base) & 85.4 & 97.3 & 83.5 & 77.8 & 81.9 & 97.0 & \textbf{91.4} & \textbf{90.1} & 96.2 & 80.7 & 81.0 & 69.8 & 91.9 \\
            RoBERTa (large) & \textbf{86.5} & 97.8 &\textbf{84.6} & 79.1 & \textbf{84.1} & 96.8 & 90.8 & 88.9 & \textbf{96.8} & \textbf{83.4} & {85.5} & 70.2 & 91.4 \\
            \midrule
            Transcormer (base)     & 84.6 & 98.1 & 80.7 & {83.2} & 80.2 & 96.0 & 90.7 & 84.1 & 95.5 & 74.3 & {85.7} & 73.4 & 91.3 \\
            {\footnotesize + BERT Init, 20K steps} & 85.0 & 98.1 & 80.0 & \textbf{84.8} & 79.2 & 96.2 & 89.3 & 84.6 & 96.1 & 76.6 & \textbf{87.8} & \textbf{75.0} & 91.7  \\
            \bottomrule

        \end{tabu}
    \end{minipage}
    \caption{Results on BLiMP. The results on GPT-2, BERT and RoBERTa are taken from \cite{Julian2020MLMScoring}. The ``+ BERT Init, 20K steps'' means the Transcormer model uses BERT model for initialization and then trains in SLM with 20K steps (nearly 5 epochs). }
    \label{table:accuracy-blimp}
\end{table*}

\subsection{Experiments on Linguistic Acceptability}
Following previous experiences~\cite{Julian2020MLMScoring}, we also conduct experiments on Benchmark of Linguistic Minimal Pairs (BLiMP)~\cite{Alex2020BLiMP}, which includes 67K minimal pairs that contrast in grammatical acceptability and isolate specific phenomena in syntax, morphology or semantics. BLiMP provides an unsupervised setting that uses language models to evaluate sentences and the acceptable sentence can be assigned by a lower log-likelihood score. We compare our model with GPT-2, BERT and RoBERTa, and the results are reported in Table~\ref{table:accuracy-blimp}. We can find that our Transcormer can easily beat GPT model in BLiMP, which further validates the effectiveness of our SLM in using bidirectional context for sentence scoring. And when compared with BERT model, our Transcormer can also match or outperform BERT performance slightly. Considering that our model only needs a single pass to produce the probability of all tokens, which also manifests the efficiency of our SLM in sentence scoring. Due to the limitation of resources, our model currently cannot beat RoBERTa since RoBERTa costs more computations than BERT and thus extract better semantics for prediction. But we believe our Transcormer can achieve the better performance by training our model with the same computations.

\section{Analyses}
\label{sec5:analysis}
In this section, we conduct some method analyses on our proposed SLM and CLM/MLM. Besides, we also provide more analyses in Appendix.

\subsection{Latency Comparisons between MLM and SLM}
To better manifest the efficiency of our SLM in rescoring when compared with MLM, we further test the inference latency between SLM and MLM at both GPU and CPU under different lengths of the input sequence. Specifically, we measure the inference latency of each model at a batch size of 1 to validate the efficiency, and all results are reported in Table~\ref{tab:latency}. From Table~\ref{tab:latency}, we can find that MLM (BERT) produces very high latency when compared with SLM (Transcormer), especially in CPU devices. Even using a smaller BERT model, it still cannot avoid the inherent issues (\ie, requires $n \times$ computations) in MLM for scoring, which needs $20 \times$ and $166 \times$ additional computations over Transcormer (SLM) in GPU and CPU. These results also demonstrate the efficiency of our model in using bidirectional context for calculating sentence scores.

\begin{table}[h!]
    \centering
    \begin{tabular}{l| l |r r | r r | r r}
    \toprule
          & &  \multicolumn{2}{c|}{\# Sent = 10} & \multicolumn{2}{c|}{\# Sent = 100} & \multicolumn{2}{c}{\# Sent = 500}\\
    Model               & \#Params & \multicolumn{1}{c}{GPU} & \multicolumn{1}{c|}{CPU} & \multicolumn{1}{c}{GPU} & \multicolumn{1}{c|}{CPU} & \multicolumn{1}{c}{GPU} & \multicolumn{1}{c}{CPU} \\
    \midrule 
    BERT (small)        & 34M  & 53ms  & 12.87s &  502ms &   317s &  3658ms & 1,650s \\
    BERT (base)         & 110M & 76ms  & 27.06s &  750ms &   703s &  7890ms & 3,210s \\
    BERT (large)        & 340M & 135ms & 59.91s & 1390ms &  1676s & 19770ms & 7,433s \\
    Transcormer (small) & 34M  &  59ms &  3.45s &   71ms &  7.64s &   183ms & 9.90s \\
    Transcormer (base)  & 110M &  67ms &  6.29s &   97ms & 17.85s & 246ms & 20.03s \\
    \bottomrule
    \end{tabular}
    \vspace{5pt}
    \caption{Inference latency between BERT (MLM) and Transcormer (SLM) at different sequence lengths. ``\# Sent'' means the length of input sequence for evaluation. GPU is evaluated at NVIDIA Tesla V100-SXM2-16GB and CPU is at Intel (R) Xeon (R) Platinum 8168 CPU @ 2.70GHz. The units of numbers in ``GPU'' and ``CPU'' columns are at millisecond (ms) and second (s).}
    \label{tab:latency}
\end{table}

\subsection{Comparable Computation between CLM and SLM}
As aforementioned, our SLM needs $3 \times$ computations when compared with CLM. To make a fair comparison, we also pre-train a Transcormer$_{small}$ with 34M parameters in total, which consists 6 transformer layers and each layer has 512 hidden size and 8 attention heads. Hence, our Transcormer$_{small}$ has approximately $\frac{1}{3}$ parameters of Transcormer$_{base}$, and has the similar computations as GPT$_{base}$. We conduct experiments on three NMT tasks and an ASR task (LibriSpeech dataset) for comparisons and the results are listed in Table~\ref{tab:nmt} and ~\ref{tab:asr} (\ie, ``Transcormer$_{small}$ row''). We can find that even under the same computation, our model still outperforms CLM and this result further validates the necessity of using bidirectional context for sentence scoring. Besides, considering that Transcormer$_{small}$ has fewer parameters, our model is also friendly to the device deployments (\eg, CPU).

\begin{table}[h]
    \centering
    \begin{tabular}{l| l | r |  c c c c}
    \toprule
    Model & Cost & PPL &  dev-clean & dev-other & test-clean & test-other \\
    \midrule
    Baseline & - & - & 2.80 & 6.90 & 3.06 & 7.05 \\
    \midrule
    MLM ($k=1$)  & $\times n$ & 4.26 & 2.30 & 5.65 & 2.59 & 5.90 \\
    MLM ($k=2$)  & $\times \left \lceil {n/2} \right \rceil$ & 8.41 & 2.41 & 5.87 & 2.70 & 6.20 \\
    MLM ($k=3$)  & $\times \left \lceil {n/3} \right \rceil$ & 11.58 & 2.60 & 5.95 & 2.87 & 6.41 \\
    MLM ($k=\frac{n}{3}$) & $\times 3$ &  - & 2.75 & 6.71 & 2.98 & 6.93 \\
    
    MLM ($k=\frac{n}{2}$)  & $\times 2$ & - & 2.80 & 6.80 & 3.01 & 6.99 \\
    \midrule
    SLM & $\times 3$ & \textbf{3.85} & \textbf{2.23} & \textbf{5.54} & \textbf{2.49} & \textbf{5.72}  \\
    \bottomrule
    \end{tabular}
    
    \vspace{5pt}
    \caption{Comparisons of sampling different $k$ tokens for prediction on MLM. We choose ASR reranking task on LibriSpeech dataset to evaluate the results and also report PPL on a subset of sentences with same length ($n=20$).}
    \label{tab:mlm}
\end{table}

\subsection{Varying Numbers of Forward Passes in MLM}
As mentioned above, MLM needs to forward $n$ passes as each time only mask one token. So what will happen if we allow each pass to mask more tokens? Therefore, we design experiments that enforces MLM to forward $k$ tokens for prediction so that it only needs $\left \lceil n/k \right \rceil$ passes, and investigate the effect of different $k$. For a certain $k$, we randomly split the sentence as $\left \lceil n/k \right \rceil$ sets, and each time masks one subset for prediction. The comparisons are listed in Table~\ref{tab:mlm}. We find that using larger $k$ will severely harm the performance, even if it can reduce the number of inference passes. When $k$ is set as $\frac{n}{3}$, which is equal to the cost of our SLM, it can hardly give any improvements over the baseline. We think that masking more tokens in the sentence will make it more difficult to estimate the token probability at a time. These comparisons also highlight the efficiency and effectiveness of our SLM  for sentence scoring. 

\begin{wrapfigure}{r}{0.4\textwidth}
    \centering
    \includegraphics[width=0.4\textwidth]{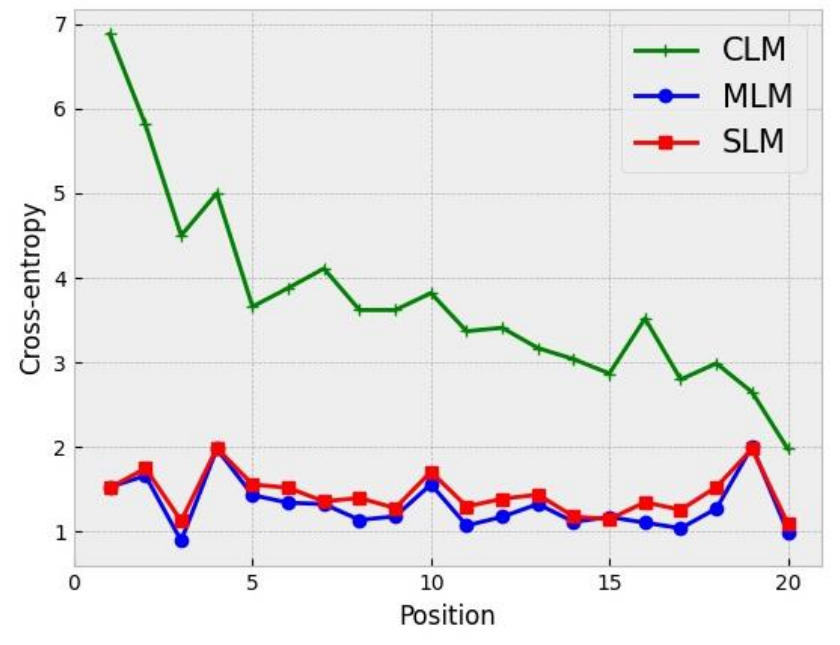}
    \caption{The average cross-entropy loss of each LM at each position. MLM uses $n$ passes to predict each position.}
    \label{fig_pos}
\end{wrapfigure}
\subsection{Sentence Scoring Quality at Each Position}
Following previous experiences~\cite{Julian2020MLMScoring}, we count the cross-entropy loss of each position in sentences for each LM, to better analyze the effectiveness of bidirectional context to estimate token probability. Specifically, we sample a subset of sentences as $\mathcal{S}$, and each sentence has the same token number $n$ (here $n$ is 20). For each position $i$, we count the average cross-entropy over the sampled subset $\mathcal{S}$ based on the output probability of each LM. The results are displayed in Figure~\ref{fig_pos}. We can find that: 1) For CLM, the cross-entropy loss is higher at the first several positions and gradually decreases for the subsequent positions but is still higher than MLM and SLM, which indicates that only using undirectional information is not enough to measure the sentence score precisely. 2) SLM can almost obtain similar loss as MLM at each position. Considering SLM just needs a single pass while MLM needs $n$ passes, this phenomenon further validates the superiority and efficiency of SLM in scoring sentences.

\section{Conclusion}
\label{sec6:conclusion}
In this paper, we propose Transcormer, a Transformer with a novel sliding language modeling for sentence scoring. Specifically, our SLM is able to produce the probability of each token over the whole sentence within a single forward pass, and  utilizes bidirectional context for prediction, and thus inherents the advantages of CLM and MLM and avoids their deficiencies. 
To the best of our knowledge, the proposed Transcormer is the first pre-trained language model tailored for sentence scoring. Experimental results on multiple datasets demonstrate the effectiveness of our Transcormer in computing sentence score for reranking tasks. 

Besides, we summarize some potential directions of our Transcormer and SLM as the future works:

\begin{itemize}[leftmargin=*]
    \item Currently our Transcormer is only conducted on the English domain under the base setting, due to limited computation. We expect to develop large-scale Transcormer and use different language domains or multilingual data for training in the future.
    \item We design sliding language modeling for sentence scoring, and our experiments are mainly on reranking task. However, based on the characteristics of our SLM, we believe our model can also be used for other scenarios (\eg, error correction~\citep{leng2021fastcorrect,leng2021fastcorrect2,song2021neural}, data selection), and we will explore the specific fine-tuning techniques when applying our SLM on different downstream tasks.  
    \item Besides, our Transcormer mainly pre-trains SLM on an encoder framework. However, our SLM is not limited to model structure. For example, SLM can be easily extended to encoder-decoder framework~\citep{song2019mass} based on paired data. Therefore, we also expect to explore the possibility of using SLM on different frameworks.
    \item Although our paper mainly focuses on text data, we want to highlight that SLM can also be extended to other different modalities with sequential characteristic (\eg, image, speech~\citep{leng2021mbnet} and time series data). Consequently, how to apply SLM to other modalities is also a valuable topic in the future.
\end{itemize}

\bibliographystyle{unsrt}
\bibliography{neurips_2022}

\section*{Checklist}

\begin{enumerate}

\item For all authors...
\begin{enumerate}
  \item Do the main claims made in the abstract and introduction accurately reflect the paper's contributions and scope?
    \answerYes
  \item Did you describe the limitations of your work?
    \answerYes{}
  \item Did you discuss any potential negative societal impacts of your work?
    \answerNA{}
  \item Have you read the ethics review guidelines and ensured that your paper conforms to them?
    \answerYes{}
\end{enumerate}

\item If you are including theoretical results...
\begin{enumerate}
  \item Did you state the full set of assumptions of all theoretical results?
    \answerNA{}
        \item Did you include complete proofs of all theoretical results?
    \answerNA{}
\end{enumerate}

\item If you ran experiments...
\begin{enumerate}
  \item Did you include the code, data, and instructions needed to reproduce the main experimental results (either in the supplemental material or as a URL)?
    \answerYes{}
  \item Did you specify all the training details (e.g., data splits, hyperparameters, how they were chosen)?
    \answerYes{}
        \item Did you report error bars (e.g., with respect to the random seed after running experiments multiple times)?
    \answerYes{Please see Section~ref{subsec:nmt} and Appendix.}
        \item Did you include the total amount of compute and the type of resources used (e.g., type of GPUs, internal cluster, or cloud provider)?
    \answerYes{Please see Section~\ref{subsec:exp_setup}}
\end{enumerate}

\item If you are using existing assets (e.g., code, data, models) or curating/releasing new assets...
\begin{enumerate}
  \item If your work uses existing assets, did you cite the creators?
    \answerYes{}
  \item Did you mention the license of the assets?
    \answerYes{}
  \item Did you include any new assets either in the supplemental material or as a URL?
    \answerNA{}
  \item Did you discuss whether and how consent was obtained from people whose data you're using/curating?
    \answerNA{}
  \item Did you discuss whether the data you are using/curating contains personally identifiable information or offensive content?
    \answerNA{}
\end{enumerate}

\item If you used crowdsourcing or conducted research with human subjects...
\begin{enumerate}
  \item Did you include the full text of instructions given to participants and screenshots, if applicable?
    \answerNA{}
  \item Did you describe any potential participant risks, with links to Institutional Review Board (IRB) approvals, if applicable?
    \answerNA{}
  \item Did you include the estimated hourly wage paid to participants and the total amount spent on participant compensation?
    \answerNA{}
\end{enumerate}

\end{enumerate}

\appendix

\section{Appendix}
\label{appendix}
\subsection{Experimental Setup}
\subsubsection{Datasets}
\textbf{IWSLT 2014} is the evaluation campaign of the 11th International Workshop on Spoken Language Translation. It consist of a lot of small-scale translation tasks collected from TED talks, including German (De), Spanish (Es), Italian (It), Dutch (NL), Polish (PL), Romanian (Ro), Russian (Ru), Turkish (Tr) to English. We randomly split each dataset as the training set and dev set with a ratio of 25:1. And each task concatenates TED.tst2010, TED.tst2011, TED.dev2010 and TED.tst2012 as the test set. The statistics of each sub-task is described as:
\begin{table}[h]
    \centering
    \begin{tabular}{l|c c c c c c c c }
    \toprule
         & De & Es & It & Nl & Pl & Ro & Ru & Tr \\
    \midrule
    Train & 160K & 169K & 167K & 153K & 128K & 167K & 153K & 109K \\
    Valid & 7.2K & 7.6K & 7.5K & 6.9K & 5.8K & 7.6K & 6.9K & 4.9K \\
    Test  & 5.5K & 5.5K & 5.5K & 5.3K & 5.4K & 5.5K & 5.5K & 5.4K \\
    \bottomrule
    \end{tabular}
    \vspace{5pt}
    \caption{Statistical of IWSLT datasets.}
    \label{tab:my_label}
\end{table}

\textbf{WMT14 English-German} comprises 4.5M bilingual data collected from Europarl v7, Common Crawl corpus and News Commentary. We concatenate newstest2012 and newstest2013 as the valid set, and choose newstest2014 as the test set for WMT14 English-German. Our experiments mainly focuses on German$\rightarrow$English.

\textbf{LibriSpeech}~\cite{Vassil2015Librispeech} includes 1000hr speech data, sampled at 16k Hz. LibriSpeech includes four subsets for evaluation, which are dev-clean, dev-other, test-clean and test-other.

\begin{table}[h!]
\small
    \centering
    \begin{tabular}{l|c | c}
    \toprule
    Hyper-parameter       & Transcormer$_{base}$ & Transcormer$_{small}$ \\
    \midrule
    Number of Layers & 12    &  6  \\
    Hidden Size      & 768   &  512 \\
    Filter Size      & 3072  & 2048 \\
    Attention heads  & 12    & 8  \\
    Dropout          & 0.1   & 0.1 \\
    Weight Decay     & 0.01  & 0.01 \\
    Learning Rate    & 5e-4 & 5e-4 \\
    Steps            & 125K  & 125K \\
    Batch            & 8192  & 8192 \\
    \bottomrule
    \end{tabular}
    \vspace{5pt}
    \caption{Pre-training hyper-parameters for Transcormer$_{base}$ and Transcormer$_{small}$.}
    \label{train_params}
\end{table}

\subsubsection{Hyper-parameter Setup}
The pre-training hyper-parameters of Transcormer are described in Table~\ref{train_params}.

\subsection{More Details about Related Works}
\label{appendix:related_works}
As mentioned in Section~\ref{sec:sentence_scoring}, some works~\cite{Alex2019BERTMouth,Julian2020MLMScoring,Kevin2020Cloze,Joongbo2020DLM} tried to alleviate the efficient problem in MLM model caused by N-passes. Specifically, \cite{Alex2019BERTMouth} proposed to calculate pseudo log-likelihood score via stochastic estimation, that is randomly sampling K tokens and computing the probability of these K tokens via masked prediction as the final sentence probability. It can reduce the time complexity from $O(|\bf{x}|)$ to $O(K)$ but will harm model performance. ~\cite{Julian2020MLMScoring} also suggested a distillation strategy to cover this problem, that requires model to compute sentence score via N-passes first (\ie, teacher model) and then distills it to the output vector of the $[$CLS$]$ token (\ie, student model) during the pre-training. However, this paradigm also will under-perform regular LMs in their experiments~\cite{Julian2020MLMScoring}. \cite{Joongbo2020DLM} designed a model named DLM to produce token-wise probability via a single inference pass. To fulfill this target, DLM only feeds word embeddings as the key/value for each Transformer layer, rather than the previous layer. Such design allows the query stream to capture the whole sentence information but without any contextualized semantics. Electric~\cite{Kevin2020Cloze} is a model built upon a Two-Cloze Tower~\cite{Alexei2019Cloze}, based on noise contrastive estimation. More in details, Electric trained a left-to-right Transformer and a right-to-left Transformer and then concatenated together at the final layer to predict each token. Just as discussed in Section~\ref{subsec:discuss}, this model learns forward and backward context individually and only fuse semantics at the final layer while the query stream of our SLM is able to fuse bidirectional context iteratively. Overall, our SLM can make full use of bidirectional information over all Transformer layers and predict token-wise probabilities simultaneously.

Besides, there also remain some works~\cite{Sumanta2021Energy,Lee2021Discriminative} which use discriminative language modeling to approximately estimate sentence scores. \cite{Sumanta2021Energy} borrowed the idea of the energy-based model into sentence reranking. \cite{Lee2021Discriminative} proposed a discriminative language model that minimizes the KL-divergence between the target distribution and the output distribution. These methods can be considered as the discriminative language modeling which directly predicts a single value to be the sentence score. Discriminative language model usually needs target datasets for fine-tuning, while our proposed language model is independent to downstream tasks. We think discriminative language models are complementary to our works and we leave this combination as the future works.

\subsection{Results}

\begin{table}[!t]
    \centering
    \begin{tabular}{l | l|c c | c c}
    \toprule
           &        &  \multicolumn{2}{c|}{Dev} & \multicolumn{2}{c}{Test} \\
    Model  & Domain &  clean & other & clean & other \\
    \midrule 
    Baseline~\cite{Joonbo2019reranking} & - & 7.17 & 19.79 & 7.26 & 20.37 \\
    \midrule 
    
    Electric~\cite{Kevin2020Cloze} & wikibooks & - & - & 5.65 & 17.42 \\
    T-TA~\cite{Joongbo2020DLM} & Libri & 4.98 & 16.09 & 5.11 & 16.91 \\
    \midrule 
    GPT-2 (117M)    & openwebtext & 5.39 & 16.81 & 5.64 & 17.60 \\
    GPT-2 (345M)    & openwebtext & 5.15 & 16.48 & 5.30 & 17.26 \\
    BERT (base)     & wikibooks   & 5.17 & 16.44 & 5.41 & 17.41 \\
    RoBERTa (base)  & RoBERTa~\cite{Yinhan2019Roberta} & 5.03 & 16.16 & 5.25 & 17.18 \\
    \midrule
    Transcormer (base) & wikibooks & 5.09 & 16.30 & 5.28 & 17.31 \\
     {\footnotesize + BERT Init, 20K steps} & wikibooks & 5.10 & 16.27 & 5.19 & 17.20 \\
    Transcormer (base, {20K steps}) & Libri & 4.61 & 15.51 & 4.73 & 16.45 \\
    \bottomrule    
    \end{tabular}
    \vspace{5pt}
    \caption{WERs on LibriSpeech after rescoring. We evaluate the results on the dataset shared by \cite{Joonbo2019reranking}, and the results of baseline, GPT and BERT are taken from \cite{Joonbo2019reranking}. RoBERTa adopts a large-scale pre-training corpus with 160GB.}
    \label{tab:acceleration}
\end{table}

\subsubsection{Comparison with other works}
As aforementioned, previous works~\cite{Joongbo2020DLM,Kevin2020Cloze} have tried some strategies to calculate the probabilities of all tokens simultaneously to avoid the limitations in N-passes. To validate the advantages of our model in using bidirectional context, we also conduct experiments to make a comparison with these methods. For the sake of fairness, all of our experiments are deployed on the same datasets used in ~\cite{Joonbo2019reranking} and the results are shown in Table~\ref{tab:acceleration}. From Table~\ref{tab:acceleration}, we find that our Transcormer can outperform these models. These improvements also demonstrate the superiority of our model in utilizing bidirectional context to predict sentence probability.

\subsection{Accelerating SLM Training}
Although SLM is more advantageous than CLM in using bidirectional context and demonstrates higher efficiency than MLM for sentence scoring, directly training SLM from scratch is still time-consuming since it needs $3 \times$ computations to maintain query and forward/backward streams. So is it possible to accelerate the training of SLM? We note that both SLM and MLM adopt the masked token plus its position to predict its target based on the context while the main difference in context is that MLM adopts one bidirectional context and SLM adopts forward and backward contexts. We think that MLM-based model with bidirectional context should also be able to produce good unidirectional context and SLM does not modify model structure, so that we can use the MLM-based model (e.g., BERT) for initialization to accelerate SLM training. Specifically, we continue to train BERT model with addition 5 epochs (nearly 20K steps) and the results can be found in Table~\ref{table:accuracy-blimp} and Table~\ref{tab:acceleration}. We can find that only needs 5 epochs, our Transcormer with BERT model initialization can match the performance that is trained from scratch, which means using BERT initialization can accelerate our SLM training.

\subsection{Analyses}
\subsubsection{Pre-training Strategy}
In our experiment setup, we use sentence-level data as the input for pre-training. To better analyze the effect of using different data processing for pre-training, we conduct experiments by using stream-level data (concatenate multiple sentences as a fixed-length, \eg, 512) to make a comparison. We apply two strategies on NMT \& ASR tasks, and then evaluate the average (top-1) accuracy of each token in a sentences with different lengths based on the output probability for SLM. The results are reported in Table~\ref{tab:data}. We can find that using stream-level data for pre-training can not achieve good accuracy when sentence length is too short. We guess that is because using stream-level data causes model can not fit short sentence since it always pre-trains under the longer sentences (\ie, 512). Considering that our downstream scenarios mainly consist of single sentence, which is usually too short, directly using stream-level data for pre-training can not achieve promising performance. As a result, we recommend to use sentence-level data for pre-training, and we also expect to explore more effective pre-training strategies in the future.

\begin{table}[h]
    \centering
    \begin{tabular}{l|c c c | c c | c c c}
    \toprule
         &  \multicolumn{2}{c}{IWSLT} & WMT & \multicolumn{2}{|c}{LibriSpeech} & \multicolumn{3}{|c}{\# Sent Len} \\
    Model & De & Es & De-En & dev-clean & dev-other & 20 & 250 & 500 \\
    \midrule
     Transcormer    & 35.24 & 41.86 & 33.10 & 2.23 & 5.54 & 60.0\% & 73.0\% & 78.8\% \\
     {\footnotesize Using stream-level} & 34.84 & 41.38 & 32.70 & 2.56 & 6.31 & 20.0\% & 55.0\% & 78.5\% \\
    \bottomrule
    \end{tabular}
    \vspace{5pt}
    \caption{Comparisons between sentence-level and stream-level pre-training. The translation direction of all IWSLT tasks is to English. We sample some sentences from wikipedia with the same length (\eg, 20, 250, 500) to evaluate the token accuracy of SLM in sentences (\ie, obtain the top-1 accuracy of each token and calculate the sentence accuracy by averaging the accuracy of all tokens).}
    \label{tab:data}
\end{table}

\subsubsection{Domain Adaption}
Following previous experiences~\cite{Julian2020MLMScoring}, we also study the effect of using in-domain data for pre-training. For NMT tasks, we randomly sample 20GB monolingual data from NewsCrawl data to build the pre-training corpus for pre-training. And for ASR tasks, as LibriSpeech includes 4GB in-domain data, we direct use this data as our pre-training corpus to handle ASR tasks. The results of NMT and ASR tasks are reported in Table~\ref{tab:domain_nmt} and Table~\ref{tab:domain_asr}. We can find that using in-domain data for pre-training is useful to improve the downstream tasks.

\begin{table}[h]
    \centering
    \begin{tabular}{l |c c c c c c c c | c}
    \toprule
              & \multicolumn{8}{c|}{IWSLT} & WMT \\
    Model     & De & Es & It & Nl & Pl & Ro & Ru & Tr & De-En \\
    \midrule
    Oracle & 41.80 & 48.69 & 41.89 & 44.38 & 27.90 & 46.01 & 29.60 & 27.25 & 39.17 \\
    \midrule
    Baseline  & 34.77 & 41.20 & 34.95 & 37.73 & 22.67 & 38.73 & 24.21 & 21.65 & 32.54 \\
    SLM (Transcormer) & {35.24} & {41.86} & 35.52 & {38.45} & {23.29} & {39.34} & {24.69} & {22.41} & {33.10} \\
    + {\footnotesize in-domain data} &  \textbf{35.74} & \textbf{42.39} & \textbf{35.97} & \textbf{39.06} & \textbf{23.91} & \textbf{39.70} & \textbf{24.95} & \textbf{23.05} & \textbf{33.51}\\
   
    \bottomrule
    \end{tabular}
    
    \vspace{5pt}
    \caption{Domain adaption on NMT tasks. The translation direction of all IWSLT tasks is to English. All results are reported in BLEU.}
    \label{tab:domain_nmt}
\end{table}

\begin{table}[h]
    \centering
    \begin{tabular}{l |c c c c }
    \toprule
    Model     & dev-clean & dev-other & test-clear & test-other \\
    \midrule 
    Oracle & 1.45 & 4.23 & 1.59 & 4.19\\
    \midrule
    Baseline  & 2.80 & 6.90 & 3.06 & 7.05 \\
    SLM (Transcormer) & {2.23} & {5.54} & {2.49} & {5.72} \\
    + {\footnotesize in-domain data} & \textbf{2.01} & \textbf{5.12} & \textbf{2.12} & \textbf{5.23} \\
    \bottomrule
    \end{tabular}
    \vspace{5pt}
    \caption{Domain adaption on LibrSpeech dataset. All results are reported in WER.}
    \label{tab:domain_asr}
\end{table}

\begin{table}[!t]
    \centering
    \begin{tabular}{l l |c}
    \toprule
    \multicolumn{2}{c|}{\textbf{Prompt Pattern}}     & \textbf{SST-2} \\
    Accepted Sentence & Unaccepted Sentence          & (zero-shot) \\
    \midrule
    {\small The sentiment of $[$\textbf{Review}$]$ is $[$\textbf{GT Label}$]$.}  & {\small The sentiment of $[$\textbf{Review}$]$ is $[$\textbf{Wrong Label}$]$.} & 58.0\% \\
    {\small $[$\textbf{Review}$]$ is a $[$\textbf{GT Label}$]$ sentiment.}  & {\small $[$\textbf{Review}$]$ is a $[$\textbf{Wrong Label}$]$ sentiment.} & 71.0\% \\
    \bottomrule
    \end{tabular}
    \vspace{5pt}
    \caption{Examples of Ranking classification for solving sentiment classification. ``[\textbf{Review}]'' means the input sentence and ``[\textbf{GT Label}]'' means the real label of the input sentence and ``[\textbf{Wrong Label}]'' means other incorrect labels. The last column is the zero-shot results on SST.}
    \label{tab:nlu}
\end{table}

\subsection{Can SLM be used for Language Understanding Tasks?}
As aforementioned in Section~\ref{subsec:discuss}, both SLM and MLM can learn bidirectional context to predict token-wise probability, so why we cannot directly use SLM for language understanding tasks? First, we want to highlight that although the query stream is able to enjoy completely bidirectional context like MLM due to the design of the triple-stream self-attention mechanism, there still remain some differences between MLM and SLM. First, the query and content streams in the MLM-based model are shared together to enjoy the benefits of the bidirectional context, while our SLM maintains query and forward/backward streams individually. In other words, the content in MLM can learn bidirectional context while SLM is used to collect forward and backward context. Considering previous state-of-the-art works~\cite{devlin2019bert,Yinhan2019Roberta,Yang2019XLNet,Zhenzhong2020Albert} have proven that BERT-style models can be fine-tuned to obtain superior performance in NLU tasks due to the benefits of bidirectional context. We think that MLM prefers NLU tasks while SLM is more suitable for scoring.

In our internal experiments, we have tried to directly fine-tune SLM on SST-2~\cite{socher2013sst} like BERT (\ie, using bidirectional context). Due to the mismatch (\ie, forward/backward context v.s. bidirectional context) between pre-training and fine-tuning, our SLM only achieves 91.9\% accuracy while the standard BERT can obtain 92.8\%. This phenomenon may also validate our hypothesis that SLM is not the optimal method for solving NLU tasks in a BERT-style fine-tuning paradigm. 

However, there still remain some alternative possibilities for SLM to adapt NLU scenarios. A possible solution is that converts language understanding tasks into ranking classification tasks. For example, assuming the task is sentiment classification and we have a review sentence, so we can construct two sentences: ``the sentiment of [review] is positive" and ``the sentiment of [review] is negative". Based on the constructed sentences, we can use our model on each sentence to calculate their sentence scores and the sentence with ground truth should have a lower log-probability score. It can be considered as a variant of prompt-based learning~\cite{Tom2020GPT3,Pengfei2021Prompt}. We have simply built two different patterns to formulate sentiment classification, and conduct experiments on SST-2 in a zero-shot setting (\ie, without fine-tuning). The results are shown in Table~\ref{tab:nlu}. We find that using patterns like ``$[$\textbf{Review}$]$ is a $[$\textbf{GT Label}$]$ sentiment.'' can obtain an accuracy of 71.0\% on SST-2 in a zero-shot setting (BERT fine-tuning cannot be used in a zero-shot scenario). These experiments also demonstrate that we can transform NLU tasks into ranking tasks and then apply our model for ranking, which also indicates the potential of our model in solving NLU or other NLP tasks. We will also continue to explore more techniques to refine this paradigm to improve performance in the future.

\end{document}